\documentclass[10pt,twocolumn,letterpaper]{article}
\pdfoutput=1 
\usepackage{iccv}
\usepackage{times}
\usepackage{epsfig}
\usepackage{graphicx}
\usepackage{amsmath}
\usepackage{amssymb}
\usepackage{widetext}
\usepackage{multirow}

\usepackage[breaklinks=true,bookmarks=false]{hyperref}

\iccvfinalcopy 

\def\iccvPaperID{****} 

\begin{document}

\title{Fuse Local and Global Semantics in Representation Learning}

\author{Yuchi Zhao \thanks{Authors contributed equally} \thanks{This work was done during the internship at Huawei Inc. } \\
University of Waterloo\\
{\tt\small y556zhao@uwaterloo.ca}
\and
Yuhao Zhou \footnotemark[1] \\
Huawei Inc.\\
{\tt\small zhouyuhao4@huawei.com}
}

\maketitle

\begin{abstract}
We propose Fuse Local and Global Semantics in Representation Learning (FLAGS) to generate richer representations. FLAGS aims at extract both global and local semantics from images to benefit various downstream tasks. It shows promising results under common linear evaluation protocol. We also conduct detection and segmentation on PASCAL\_VOC \cite{pascal} and COCO \cite{coco} to show the representations extracted by FLAGS are transferable.
\end{abstract}

\section{Introduction}
Self-supervised contrastive learning has been shown great potential in extracting general visual  representations recently \cite{chen2020simple}. Models pre-trained in such a manner achieve better results in downstream tasks like object detection and segmentation, comparing to fully-supervised pre-trained models \cite{he2020momentum}. Besides, the gap of performance in the basic image classification is also narrowing. However, problems with self-supervised learning are discovered recently, including inductive bias and sensitivity to data augmentation. These are mainly caused by the weak-supervising signal and image pairs used in the contrastive mechanism. The goal of learning general representations has not been achieved yet. 

We propose a novel framework FLAGS  with a custom sampling strategy that incorporate features from different latent spaces. Our motivation is to combine the strengths of unsupervised contrastive learning and fully-supervised learning to generate richer representations. We argue that there are two levels of semantics in images, global semantics and local semantics. Global semantics refer to all entities in the image including the background and the foreground. Local semantics refer to the entity that is defined by the label of the image. Self-supervised contrastive learning methods perform instance discrimination which can be seen as learning global semantics. While the supervised learning methods on the other side, learn the local semantics. As shown in  Table 1, We find a trend that tasks like object detection and segmentation benefit from global semantics, whereas image classification benefits more from local semantics. In details, for image classification, the supervised method which learns local semantics achieves much better accuracy than self-supervised methods which learn global semantics. On the other hand, for detection and segmentation which require more information besides the semantics of the entity of interest, self-supervised methods perform close to or even better than the supervised method. Thus intuitively a method learns both global and local semantics should be a better pre-training for larger variety of tasks, performing better than the methods learn either one. The authors in \cite{wei2020semantic} also argue that self-supervised methods and fully-supervised methods learn different kinds of representations. They try to alleviate the conflict between the fully-supervised and unsupervised learning objectives. Comparing to their method, our method aims to learn different level of semantics features thoroughly with equal attention on global and local semantics. Our contributions are summarized below:

\begin{enumerate}
\item	We design a novel architecture which can effectively guide feature extractors to learn both global and local semantic features. 
\item	We avoid the conflict between the learning objectives, by using a sampling strategy and splitting the learning of two semantic features by projecting them into separate hyperspheres.
\item	Our method of guiding contrastive networks opens a new direction of research for general representation learning. 
\end{enumerate}

\begin{table*}
\begin{center}
\centering
\label{Table_1}
\caption{ Comparison of different pre-training methods on three downstream tasks. The standard backbone is ResNet-50. Linear evaluation protocol \cite{he2020momentum} is applied for image classification. For object detection and semantic segmentation, models are fine-tuned on the PASCAL VOC dataset \cite{pascal}.}
\begin{tabular}{c|c|c|c|c}
    \hline
    Type of semantics &pre-train &  Classification (Accuracy\%) &  Detection ($AP_{50}$)  & Segmentation (mIOU)  \\ \hline
    \multirow{4}{*}{global} &Colorization \cite{zhang2016colorful}  & 39.6    &-       &-   \\ \cline{2-5}
    & Jigsaw \cite{noroozi2017unsupervised}  & 44.6    &61.4    &-   \\ \cline{2-5}
     &SimCLR \cite{chen2020simple}           & 69.3    & 79.4 \cite{wang2021dense}  & 64.3 \cite{wang2021dense}   \\ \cline{2-5}
     &MoCo-v2 \cite{chen2020improved}        & 71.1    & 82.5       & 67.5 \cite{wang2021dense}  \\ \hline
    local &supervised \cite{he2015deep}           & 76.5    & 81.3       & 67.7 \cite{wang2021dense}    \\ \hline
\end{tabular}
\end{center} 
\end{table*}

\section{Related Work}
In this section, we summarize the current development in representation learning. The objective of representation learning is to abstract and disentangle underlying factors of variation existing in raw data  \cite{bengio2014representation}. A robust representation contains unique properties that can be easily distinguished from others. To achieve this goal, different learning strategies are explored. 

\subsection{Supervised Learning}
Fully-supervised methods heavily rely on labels as strong supervising signals when training. The objective is to learn a representation that maximizes the probability of finding the correct label of images. In terms of network architectures, different convolutional neural networks \cite{simonyan2015deep, he2015deep, tan2020efficientnet} are commonly used to encode input images and generate feature maps for classification and other tasks. Recently, transformer-based models have been shown to better encode visual information and achieved promising results in downstream tasks \cite{dosovitskiy2021image, carion2020endtoend, zheng2021rethinking}. Commonly, these methods train on large datasets such as ImageNet \cite{imagenet_cvpr09} and JFT-300M \cite{sun2017revisiting} which capture a great variation of objects. The extracted representations primarily encode specific features biased to the learning objective. Geirhos et al. \cite{geirhos2019imagenettrained} has shown through experiments that CNN models trained on ImageNet are biased to texture features that are beneficial to object recognition tasks. This indicates the learned representation only captures partial invariance of objects.

\subsection{Pretext Learning}
Learning to perform a predefined pretext task is a discriminative approach in self-supervised learning. Pretext tasks are hand-designed to learn desired features using self-generated pseudo labels. Specifically, image inpainting \cite{pathak2016context} and colorization \cite{zhang2016colorful} are generation-based tasks where the objective is to guide models to recover incomplete images. To achieve these tasks, both semantic and context information of scenes are necessary to be learned. Solving image jigsaw puzzle \cite{noroozi2017unsupervised, kim2018learning} and image rotation \cite{gidaris2018unsupervised} guide networks to encode positional and global context information of images. However, experiments from Chen et al. \cite{chen2020simple} have demonstrated the gap between the model performance on downstream tasks trained using pretext task methods and fully-supervised methods as shown in Table 1. The common problem is that a single pretext task cannot provide enough supervision signal when training and it limits networks to learn biased representations. 

\subsection{Contrastive Learning}
Contrastive learning methods utilize instance discrimination as the pretext task and it shows the potential to better extract useful representation that transfers well in downstream tasks. In the beginning, image features are extracted by an encoder and they are projected into a latent space. Contrastive loss \cite{1640964} is then used to measure the similarity(distance) between those projected features. For supervised contrastive learning,  SCL \cite{khosla2021supervised} utilizes images with the same label as positive samples to contrast which achieves competitive results as traditional supervised methods. In unsupervised contrastive learning,  MoCo \cite{he2020momentum} introduces a dynamic dictionary with the momentum update mechanism which extracts representation by comparing with a diverse set of negative samples. SimCLR \cite{chen2020simple} proves the importance of data augmentation and large batch size in a simple contrastive setting. However, data augmentation introduces inductive bias that restricts the generalization of representations. Experiments show that both strength and selection of data augmentation affect the performance of pre-trained models. As shown by Purushwalkam and Gupta \cite{purushwalkam2020demystifying}, data augmentations like random cropping learn occlusion invariance. Besides, instead of contrasting with different images within the same class, such approaches do not use class labels and can only contrast with augmented images of themselves which fail to explore object invariance. 

\begin{figure*}
\begin{center}
\includegraphics[width=0.64\linewidth]{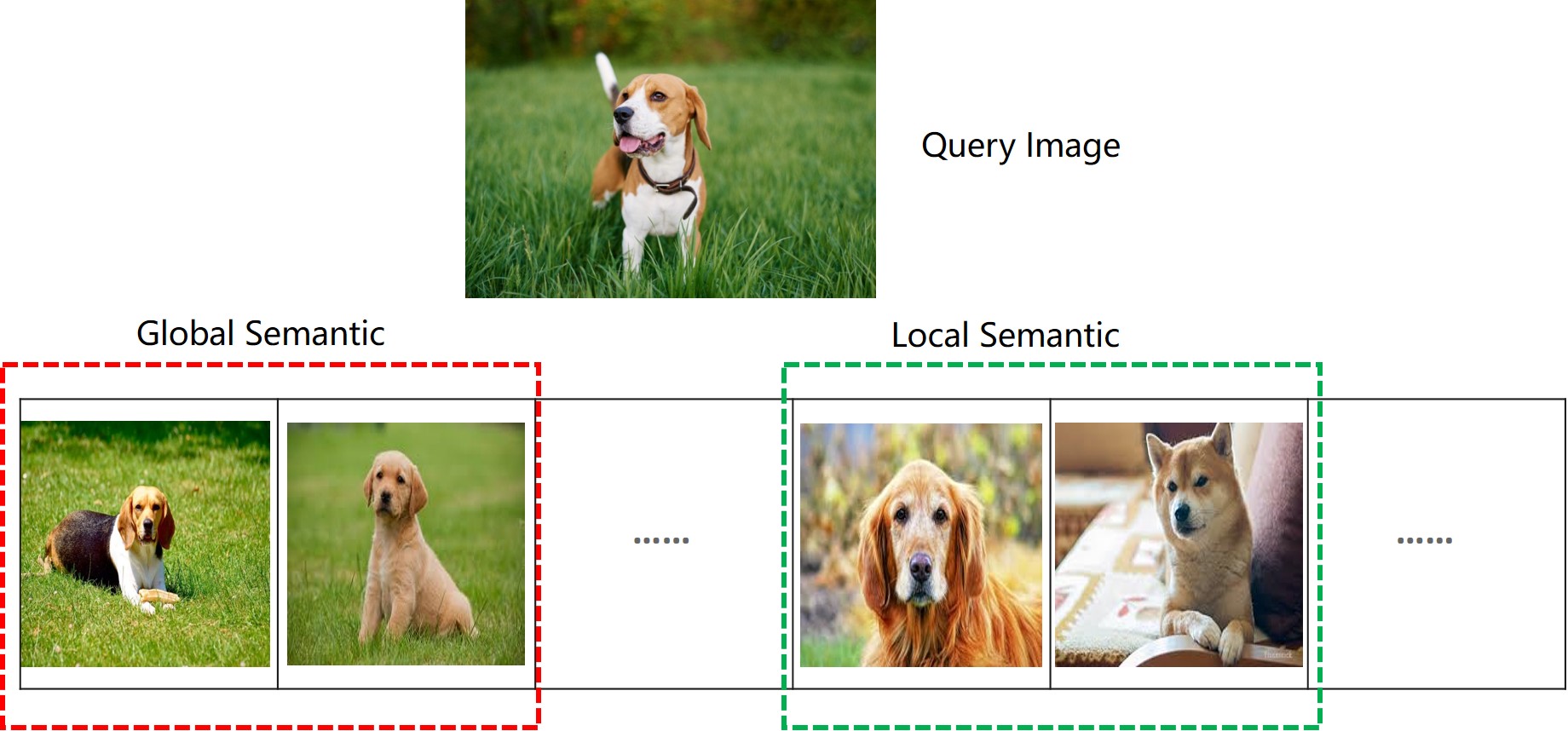}
\end{center}
   \caption{Illustration of the sorted images in the same class based on cosine similarity. 
   In this example, the top-2 images are selected as global positive pairs since they have similar background as the query image. While the two images in the middle of the list, do not have the similar background as the query image. So they are selected as local positive pairs.}
   
   
\label{fig:short}
\end{figure*}

\section{Method}
\subsection{Preliminary Knowledge}
\textbf{Supervision Signals} To learn a desired representation, specific supervision signals need to be supplied to guide networks during training. However, direct combination of supervision signals injects confusion into learning objectives. In self-supervised contrastive learning, methods are sensitive to data augmentation which serves as a weak supervision signal. To tackle this problem, LooC \cite{xiao2021contrastive} projects features into different latent subspaces for each augmentation so the invariance of raw data and the variance of augmentation are both preserved in the extracted representations. We argue that this concept of disentangling features into separate latent spaces can help to learn a more robust representation with stronger guidance.  

\textbf{Adjustment In Positive Pair Selection} The selection of positive pairs plays an important role in contrastive learning. In self-supervised settings, augmented images of themselves are treated as positive sample whereas other images including those from the same classes are treated as negatives. In supervised settings, SCL \cite{khosla2021supervised} treats all samples from the same class as positives and other remaining samples in the batch as negative samples, which networks learn a richer representation within local regions. Recent development of SCAN \cite{wei2020semantic} first reveals the conflicting objectives in supervised and self-supervised learning and shows task-agnostic appearance information are learned in common self-supervised methods. To alleviate such conflict in contrastive learning, SCAN selects the top-k nearest neighbors of query image from the same class as positives in the feature space formulated by pre-trained MoCo-v2, which the pre-trained model has been proved to generate more robust representations. 

\begin{figure}[t]
\label{figure_2}
\begin{center}
\includegraphics[width=0.97\linewidth]{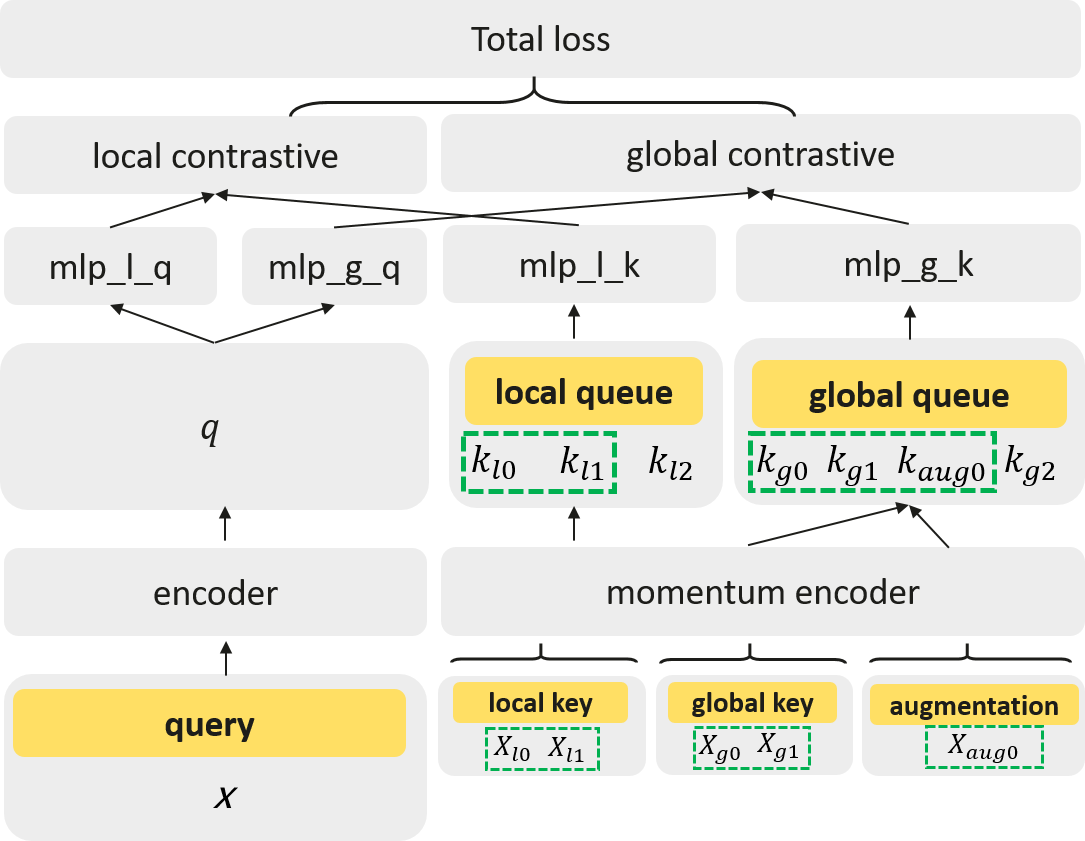}
\end{center}
   \caption{\textbf{The main architecture of FLAGS} At each training step, a query image X is processed by the encoder. A local key pair \{$X_{l0}, X_{l1}$\}, a global key pair \{$X_{g0}, X_{g1}$\}, and an augmented version of query $X_{aug0}$ are fed into the momentum encoder. The output features \{$k_{l0}, k_{l1}, k_{g0}, k_{g1}, k_{aug0}$\} are then projected to corresponding local and global subspace via mlp\_l\_k and mlp\_g\_k. The encoded query q is projected into local and global subspace via mlp\_l\_q and mlp\_g\_q. Losses for each subspace are calculated separately and combined to form the total loss as shown in the loss function section. Two queues are maintained for local and global semantic pairs. }
\label{fig:long}
\label{fig:onecol}
\end{figure}

\subsection{Our Approach}
Our goal is to enable networks to learn rich representations that can smoothly adapt to various downstream tasks. We first define that self-supervised contrastive methods learn \textbf{global semantics} and fully-supervised methods learn \textbf{local semantics}. Specifically, global semantics encode rough context and semantic information of all objects throughout the region. Local semantics focus on extracting detailed representations of specific objects located at certain regions. Our intuitive is to guide models to encode both local and global semantics by fusing the learning objectives of fully-supervised learning and contrastive learning. Inspired by LooC \cite{xiao2021contrastive}, we propose a novel supervised contrastive framework FLAGS as shown in Figure 2. The core idea of FLAGS is to project pairs of features into either global or local subspace. Specifically, a query image with a local and a global positive pair (keys) are fed into the encoders, then the output features are projected into corresponding subspaces using two separate sets of MLPs. We maintain two queues to store local and global keys that are used to contrast during training. Losses are computed independently for each subspace and combined in the end. 

\newpage
\begin{widetext}
\begin{equation}
\label{equation 1}
\mathcal{L}_{ALL} = \sum_{i\in I} \mathcal{L}_{global}(i) + \sum_{i\in I}\mathcal{L}_{local}(i) 
\end{equation}

\begin{equation}
\label{equation 2}
\mathcal{L}_{global}(i) =  \frac{-1}{|P_g(i)|} 
                \sum_{p\in P_g(i)} 
                    \log \frac{\exp(z_g(i) \cdot z_g(p) / \tau )}
                             {\sum\limits_{a\in Q_g}\exp(z_g(i) \cdot z_g(a) / \tau) + {\exp(z_g(i) \cdot z_g(p) / \tau )}}
\end{equation}
\begin{equation}
\label{equation 3}
\mathcal{L}_{local}(i) = \frac{-1}{|P_l(i)|} 
                \sum_{p\in P_l(i)} 
                    \log \, 
                        \frac{\exp(z_l(i) \cdot z_l(p) / \tau )}
                             {\sum\limits_{a\in Q_l}\exp(z_l(i) \cdot z_l(a) / {\tau)+\exp(z_l(i) \cdot z_l(p) / \tau )}}
\end{equation}
\end{widetext}

\textbf{Image Pairs Generation} To form image pairs for each query image that will be fed into different subspaces, we adapt the image pair selection strategy in SCAN \cite{wei2020semantic} and modify it accordingly. In details, images are fed into the ResNet pre-trained in MoCo-v2 to generate features \(p\) with the size of [1, 1024]. The cosine similarity between each image within the same class is calculated. Then images are sorted using the similarity and stored into a list  as shown in Figure 1. We select the top-2 similar images as a pair to project on global-semantic subspace. To alleviate the conflict between global and local learning objectives,  two images at middle of lists are chosen to form a pair for local-semantic subspace. Since their extent of global semantics similarity with the query image is moderate. 

\textbf{Loss Function} 
Our loss function is based on the common contrastive loss \cite{1640964}. We call it combined contrastive loss \autoref{equation 1}. It is the summation of contrastive loss at the global branch and the contrastive loss at the local branch. \autoref{equation 2} shows the loss for one query image at the global branch. \autoref{equation 3} shows the loss for one query image at the local branch. The form of equation is inspired by the supervised contrastive loss introduced in SCL \cite{khosla2021supervised}. It allows the contrastive loss function to generalize to an arbitrary number of positives. In details, there are $N$ query images in one batch. Let $i$ be the index of images in the batch. For the $i$-th query image, it has global positive keys and its augmentation $P_g(i)$. It also has local positive keys $P_l(i)$. $P_g(i)$ has dimension $M+1$ where $M$ is the number of positive keys and one represents the augmentation. $P_l(i)$ has dimension $M$. $Q_g$ is the queue that contains negative images for the global semantic branch and $Q_l$ is the queue for the local semantic branch. $Q_g$ and $Q_l$ both has dimension of $K$. $Z_g(i)$ is the normalized projected features of image $i$ in the global subspace. Accordingly, $Z_l(i)$ is in the local subspace. $\tau \in \mathbb R$ is temperature parameter. 

\section{Experiments}
\subsection{Pre-training}
\textbf{Dataset Preparation} 
We use ImageNet-1M \cite{imagenet_cvpr09} for pre-training and validation. We use the proposed image pair generation strategy to prepare the global and local positive pairs for each image in the ImageNet-1M \cite{imagenet_cvpr09} train set. These positive pairs are keys for the corresponding query image during training.

\textbf{Training} 	
We perform three different pre-training methods. The first pre-training method is FLAGS with only the global branch. The second pre-training method is FLAGS with both the global and the local branch. The last is MoCo\_v2 \cite{chen2020improved} which using the same loss as FLAGS. The hyper parameters are basically the same as those used in MoCo \cite{he2020momentum}. We use SGD as the optimizer with a momentum of 0.9. The learning rate is 0.03. The batch size is 256. We use the loss function shown in Section 3.2. The checkpoint at the 200th epoch of each pre-training is used for the rest of experiments. Our train set contains about 1.23 million images rather than normal 1.28 million images. 50,000 images are evenly taken out of the train set from each class. 

\subsection{Linear Evaluation}
We follow the common linear evaluation protocol where we freeze the weights of the pre-trained model except the fully-connected layers. Then, we train a supervised classifier using the ImageNet-1M \cite{imagenet_cvpr09}. The hyper parameters are the same as those used in MoCo \cite{he2020momentum}. The top-1 accuracy is recorded during training after each epoch and the top accuracy is shown in  Table 2. As we can see, FLAGS models achieve much higher accuracy than MoCo \cite{he2020momentum}. This is expected because FLAGS models used more supervised signals during pre-training. At the same time, FLAGS with a local branch has a lower accuracy than FLAGS without a local branch. This might be due to the noise added when we select positive local pairs. The selection strategy for local positive pair should vary for different query images or different classes. Since currently, the local key might be too different from the global key which creates a conflicting signal that makes the model confused. In other words, we tell the model to encode two very different images in the same way. This may explains the accuracy difference between FLAGS with and without the local branch. 

\begin{table}[t]
\begin{center}
\centering
\label{Table_2}
\caption{Linear evaluation accuracy of ResNet models trained using different methods}
\begin{tabular}{c|c}
    \hline
    $\ $ & Accuracy\%\\
    \hline
    MoCo                   & 67.16  \\ \hline
    FLAGS aug+global        & 78.45  \\ \hline
    FLAGS aug+global+local  & 77.75  \\ 
    \hline
\end{tabular}
\end{center} 
\end{table}

\subsection{Object Detection and Segmentation}
We use object detection and segmentation as downstream tasks to evaluate how good the features are transferable. We perform the experiments with two datasets, PASCAL\_VOC\_2007 \cite{pascal} and COCO\_2017 \cite{coco}. In all experiments, we train detectors with the ResNet R50-C4 \cite{he2015deep} as the feature encoder, where the weights of encoder are initialized from the three pre-trained models, two of them using FLAGS and one using MoCo. We then fine-tune the whole detectors end-to-end. The same hyper parameters settings are used for all experiments.  

\textbf{COCO Object Detection and Instance Segmentation} 	
We use Mask R-CNN \cite{mask} as the detector with the backbone of ResNet R50-C4 \cite{he2015deep}. The batch size is 8 and iteration is 180,000. These mean that the total epoch is about the same as 1*schedule defined in detectron2 \cite{detectron2}. The learning rate is 0.01. We fine-tune models using COCO train\_2017 and evaluate using COCO val\_2017. The results are shown in  Table 3 and  4. As we can see, initialized with weights of FLAGS with global and local branch surpass with weights of a supervised pre-training in both detection and segmentation. This shows the effectiveness of our proposed method. 

\begin{table}[t]
\begin{center}
\centering
\label{Table_3}
\caption{Object detection results of Mask R-CNN on COCO dataset with different pre-trained ResNet R50-C4 models.}
\begin{tabular}{c|c|c|c}
    \hline
    pre-train                & AP        & $AP_{50}$  & $AP_{75}$  \\ \hline
    supervised 1*schedule    & 38.200    & 58.200     & 41.200     \\ \hline
    MoCo batch size 8        & 37.879    & 57.201     & 40.903     \\ \hline
    FLAGS aug+global          & 38.344    & 57.852     & 41.357     \\ \hline
    FLAGS aug+global + local  & 38.197    & 58.053     & 40.994     \\ 
    \hline
\end{tabular}
\end{center} 
\end{table}

\begin{table}[t]
\begin{center}
\centering
\label{Table_4}
\caption{Instance segmentation results of Mask R-CNN on COCO dataset with different pre-trained ResNet R50-C4 models.}
\begin{tabular}{c|c|c|c}
    \hline
    pre-train                  & AP          & $AP_{50}$    & $AP_{75}$  \\ \hline
    supervised 1*schedule      & 33.300      & 54.700       & 35.200     \\ \hline
    MoCo batch size 8          & 33.312      & 54.055       & 35.598     \\ \hline
    FLAGS aug + global          & 33.511      & 54.461       & 35.505     \\ \hline
    FLAGS aug + global + local  & 33.431      & 54.552       & 35.460     \\ 
    \hline
\end{tabular}
\end{center} 
\end{table}

\textbf{PASCAL VOC Object Detection} 
The detector used is Faster R-CNN \cite{faster} with R50-C4 \cite{he2015deep} as backbone. The batch size is 4 and the number of iteration is 96,000. The learning rate is 0.005. The same feature normalization (sync norm) described in MoCo \cite{he2020momentum} is used. Other parameters are defaults defined in detectron2 \cite{detectron2}. We fine-tune models using VOC trainval\_2012 and VOC trainval\_2007. Evaluation is conducted on VOC test\_2007. The results are shown in Table 5. FLAGS with only global branch gets similar performance as MoCo \cite{he2020momentum}. When the local branch is added, the performance decreases. This is opposite to the results with COCO \cite{coco} where FLAGS with local branch achieves the best overall scores. We think this phenomenon is due to the difference between two datasets. As survey shows, in PASCAL\_VOC \cite{pascal} more than 40\% instances take above 50\% of whole image size. On the other side, in COCO \cite{coco} only 1\% instances take more than 50\% of whole image size. In addition, COCO \cite{coco} has 7.3 objects per image and PASCAL\_VOC \cite{pascal}  has 2.3 objects per image. These differences mean that detection and segmentation requires more local semantics to get into details of part of images in COCO \cite{coco}. FLAGS’s local branch successfully captures local semantics, improving the detector’s performance on COCO \cite{coco}. 

\begin{table}[t]
\begin{center}
\centering
\label{Table_5}
\caption{Object detection results of Mask R-CNN on  PASCAL\_VOC dataset with different pre-trained ResNet R50-C4 models.}
\begin{tabular}{c|c|c|c}
    \hline
    pre-train               & AP      & $AP_{50}$  & $AP_{75}$  \\ \hline
    MoCo                    & 50.571  & 78.879     & 55.193     \\ \hline
    FLAGS aug+global         & 50.701  & 79.328     & 54.310     \\ \hline
    FLAGS aug+global+local   & 48.141  & 78.348     & 51.034     \\ 
    \hline
\end{tabular}
\end{center} 
\end{table}

\section{Conclusion}
In this paper, we propose FLAGS that extracts rich and transferable representations for various downstream tasks. We suggest representations have two levels of semantics: global semantics and local semantics. By contrasting with local and global image pairs in different subspaces, models benefit from both learning objectives. Through a few experiments, we demonstrate the learned representations based on FLAGS improves the performance in image classification, object detection and segmentation. It has been shown that this direction is promising and we hope others can expand the research. The positive pair sampling strategy can be optimized to get more precise keys. In addition, more visualizations of learned representations are beneficial.

\newpage
\section*{Acknowledgement}
We would like to express our great appreciation to Dr. Hailin Hu for providing tremendous support and constructive suggestions. We also want to thank Xiangqian Wang and Lin Du for facilitating the publishing. Last but not least, thanks Ce Wang for the great effort in validating experiment results.

{\small
\bibliographystyle{ieee}
\bibliography{ms}
}

\end{document}